\documentclass[10pt,twocolumn,letterpaper]{article}

\usepackage{cvpr}
\usepackage{times}
\usepackage{epsfig}
\usepackage{graphicx}
\usepackage{amsmath}
\usepackage{amssymb}
\usepackage[numbers,sort&compress]{natbib}

\usepackage[pagebackref=true,breaklinks=true,letterpaper=true,colorlinks,bookmarks=false]{hyperref}

\cvprfinalcopy % *** Uncomment this line for the final submission

\def\cvprPaperID{9} % *** Enter the CVPR Paper ID here

\ifcvprfinal\fi
\begin{document}

%%%%%%%%% TITLE
\title{The iWildCam 2020 Competition Dataset}

\maketitle

\begin{abstract}
   Camera traps enable the automatic collection of large quantities of image data. Biologists all over the world use camera traps to monitor animal populations. We have recently been making strides towards automatic species classification in camera trap images. However, as we try to expand the geographic scope of these models we are faced with an interesting question: how do we train models that perform well on new (unseen during training) camera trap locations? Can we leverage data from other modalities, such as citizen science data and remote sensing data? In order to tackle this problem, we have prepared a challenge where the training data and test data are from different cameras spread across the globe. For each camera, we provide a series of remote sensing imagery that is tied to the location of the camera. We also provide citizen science imagery from the set of species seen in our data. The challenge is to correctly classify species in the test camera traps.
\end{abstract}

\section{Introduction}
 In order to understand the effects of pollution, exploitation, urbanization, global warming, and conservation policy on our planet's biodiversity, we need access to accurate, consistent biodiversity measurements. 
 Researchers often use \emph{camera traps} -- static, motion-triggered cameras placed in the wild -- to study changes in species diversity, population density, and behavioral patterns. 
 These cameras can take thousands of images per day, and the time it takes for human experts to identify species in the data is a major bottleneck.
 By automating this process, we can provide an important tool for scalable biodiversity assessment. 
 
\begin{figure}
\begin{minipage}[b]{.3\linewidth}
  \centering
  \centerline{\includegraphics[width=3cm]{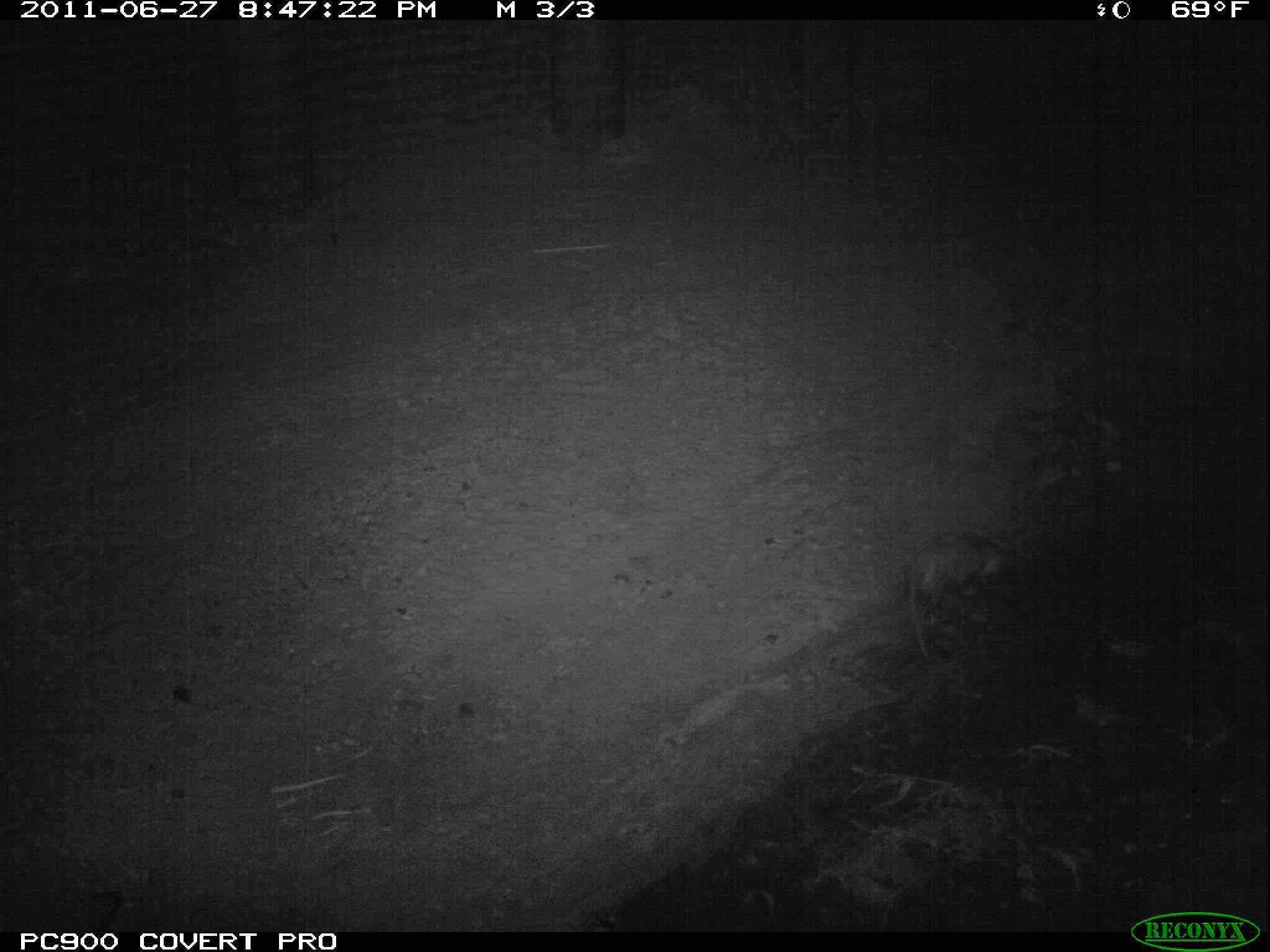}}
  \vspace{.05cm}
  \centerline{(1) Illumination}\medskip
\end{minipage}
\hfill
\begin{minipage}[b]{0.3\linewidth}
  \centering
  \centerline{\includegraphics[width=3cm]{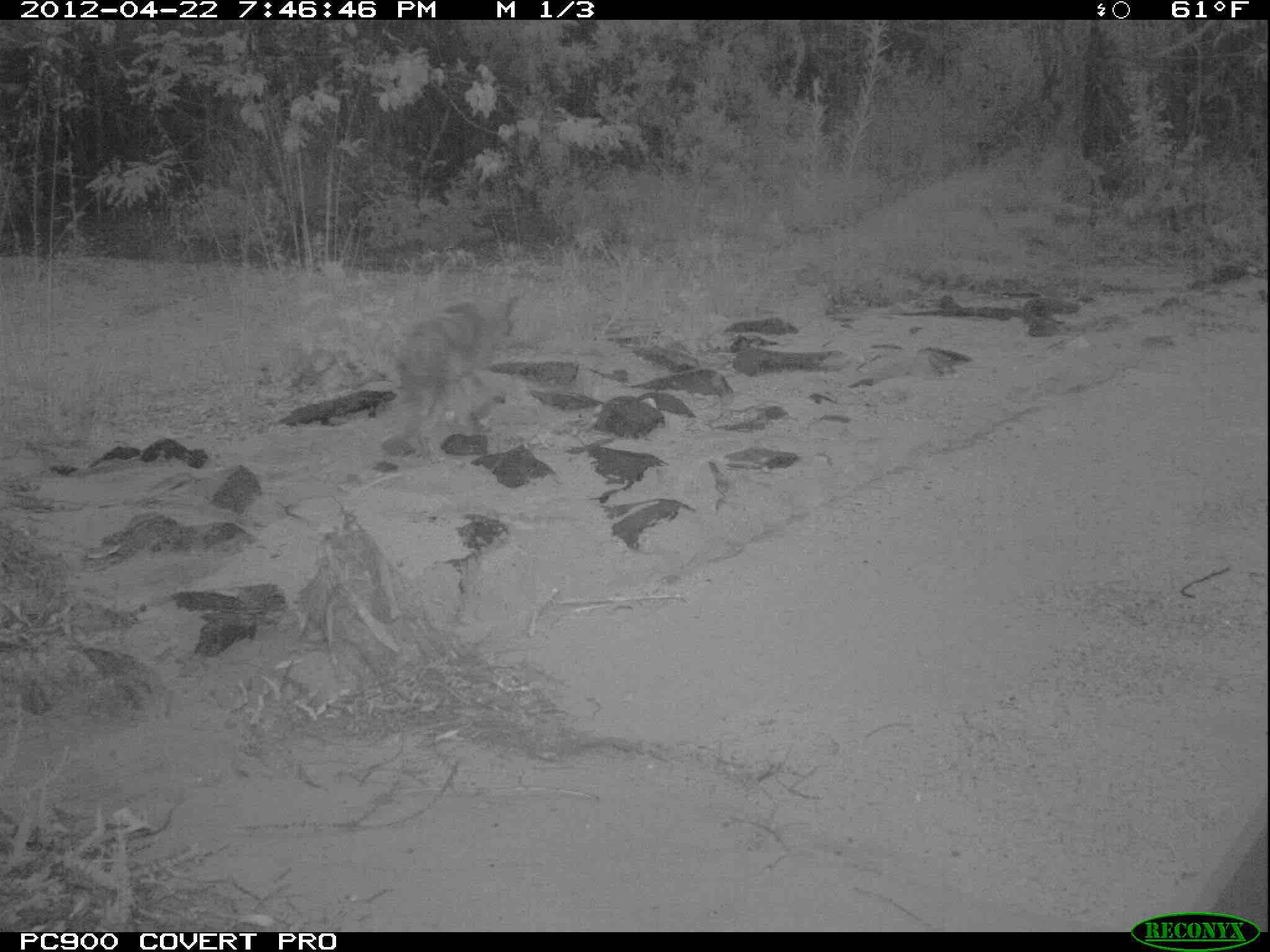}}
  \vspace{.05cm}
  \centerline{(2) Blur}\medskip
\end{minipage}
\hfill
\begin{minipage}[b]{.3\linewidth}
  \centering
  \centerline{\includegraphics[width=3cm]{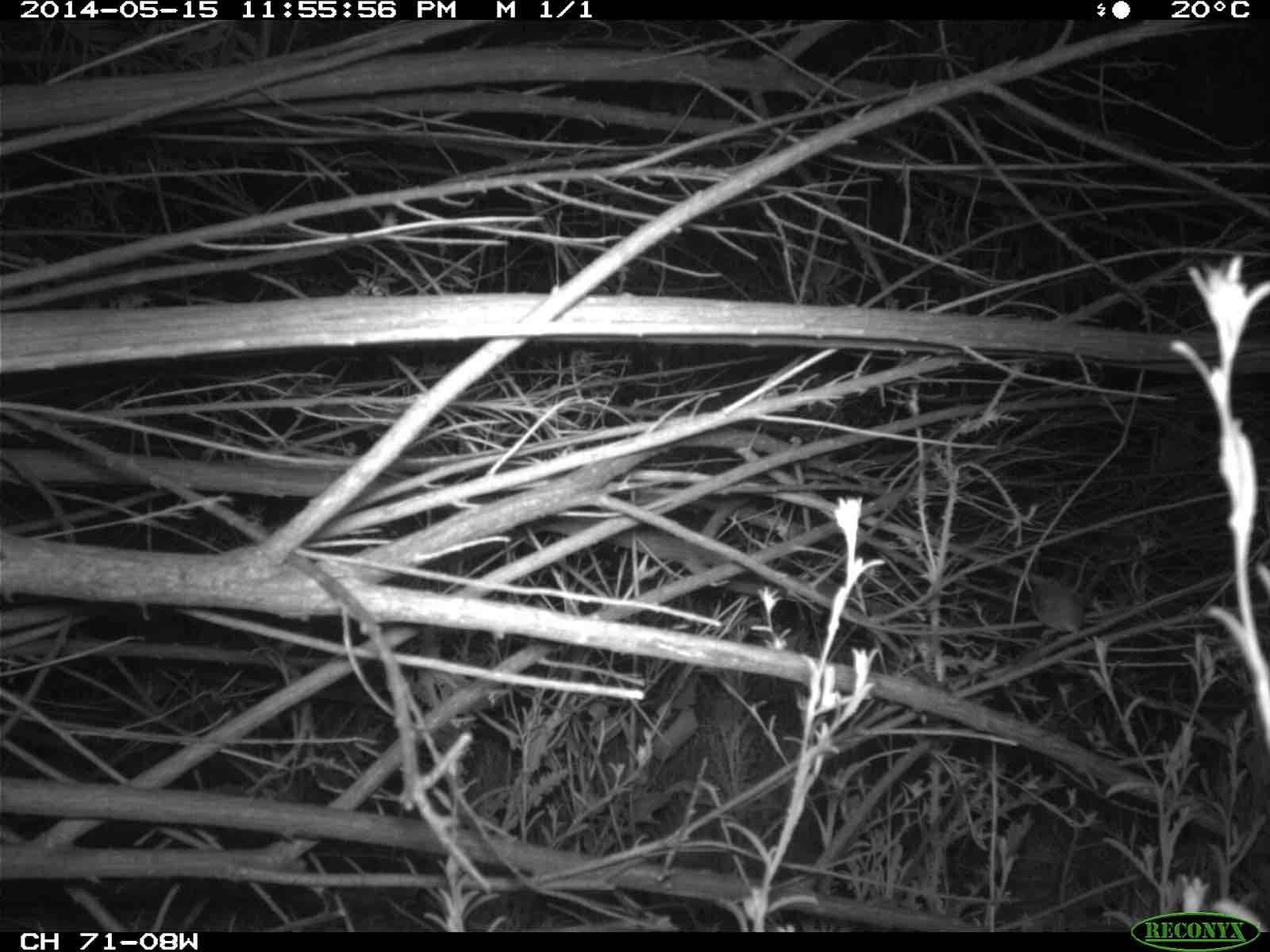}}
  \vspace{.05cm}
  \centerline{(3) ROI Size}\medskip
 \end{minipage}
\begin{minipage}[b]{0.3\linewidth}
  \centering
  \centerline{\includegraphics[width=3cm]{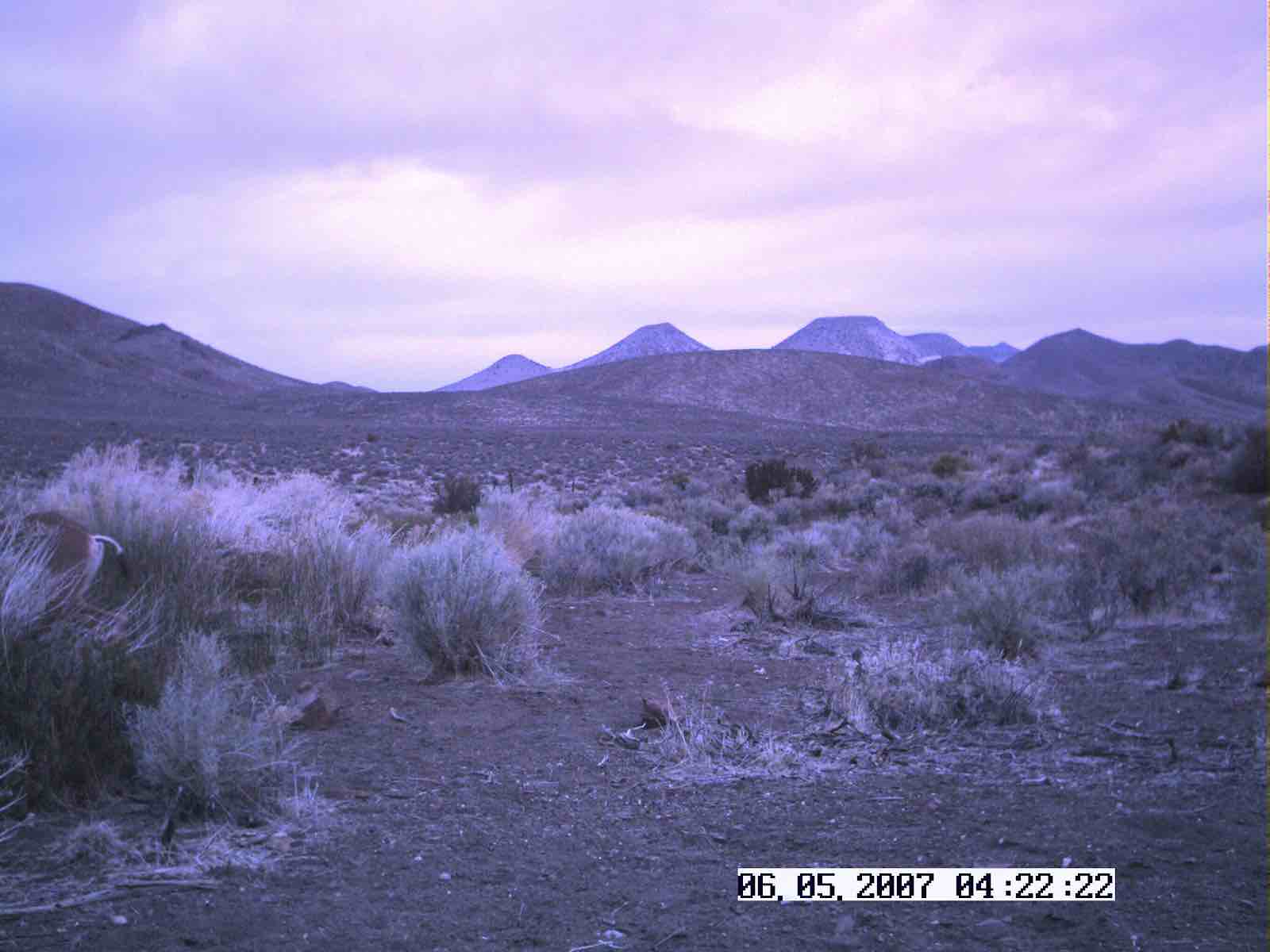}}
%  \vspace{1.5cm}
  \centerline{(4) Occlusion}\medskip
\end{minipage}
\hfill
\begin{minipage}[b]{.3\linewidth}
  \centering
  \centerline{\includegraphics[width=3cm]{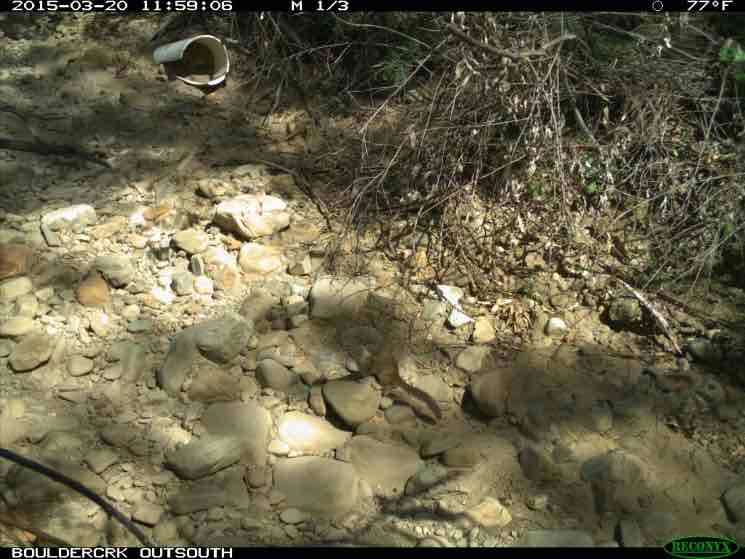}}
%  \vspace{1.5cm}
  \centerline{(5) Camouflage}\medskip
\end{minipage}
\hfill
\begin{minipage}[b]{0.3\linewidth}
  \centering
  \centerline{\includegraphics[width=3cm]{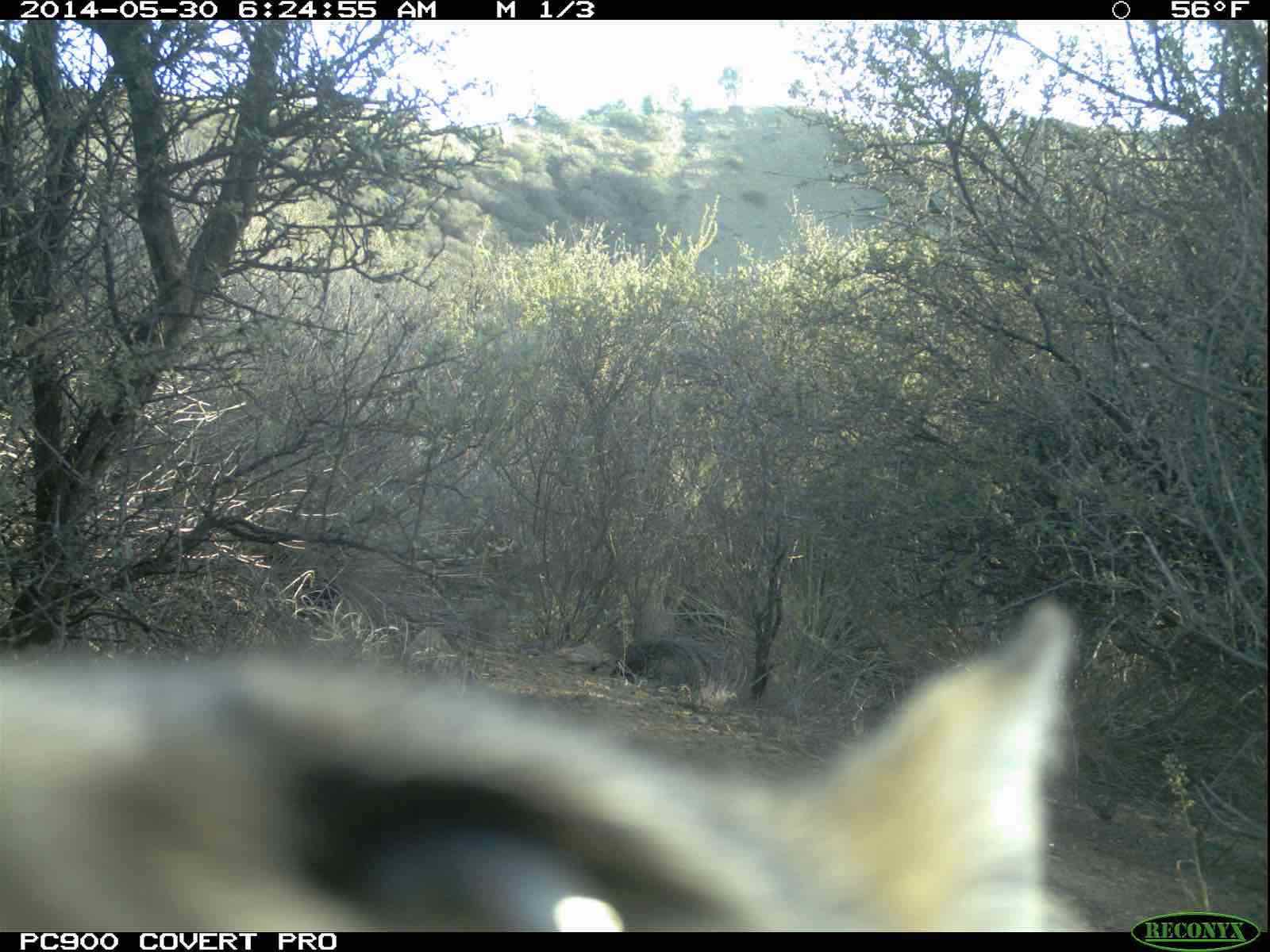}}
%  \vspace{1.5cm}
  \centerline{(6) Perspective}\medskip
\end{minipage}
\caption{\textbf{Common data challenges in camera trap images.} (1) {\bf Illumination}: Animals are not always well-lit. (2) {\bf Motion blur}: common with poor illumination at night. (3) {\bf Size of the region of interest} (ROI): Animals can be small or far from the camera. (4) {\bf Occlusion}: e.g. by bushes or rocks. (5) {\bf Camouflage}: decreases saliency in animals' natural habitat. (6) {\bf Perspective}: Animals can be close to the camera, resulting in partial views of the body.}
\label{fig:challenging_ims}
\end{figure}

 Camera trap images are taken automatically based on a triggered sensor, so there is no guarantee that the animal will be centered, focused, well-lit, or at an appropriate scale (they can be either very close or very far from the camera, each causing its own problems). 
 See Fig. \ref{fig:challenging_ims} for examples of these challenges. 
 Further, up to 70\% of the photos at any given location may be triggered by something other than an animal, such as wind in the trees. 
Automating camera trap labeling is not a new challenge for the computer vision community \cite{wilber2013animal,chen2014deep,zhang2016animal,miguel2016finding,giraldo2017camera,yousif2017fast,villa2017towards,norouzzadeh2017automatically, beery2018recognition, beery2020synthetic,beery2019long, norouzzadeh2019deep, schneider2018deep, beery2019efficient, tabak2020improving, norouzzadeh2019deep}. 
However, most of the proposed solutions have used the same camera locations for both training and testing the performance of an automated system. 
If we wish to build systems that are trained to detect and classify animals and then deployed to new locations without further training, we must measure the ability of machine learning and computer vision to \emph{generalize to new environments}~\cite{beery2018iwildcam, tabak2020improving}. 
This is central to the 2018 \cite{beery2018iwildcam}, 2019 \cite{beery2019iwildcam}, and 2020 iWildCam challenges. 

\begin{figure}
    \centering
    \includegraphics[width=0.45\textwidth]{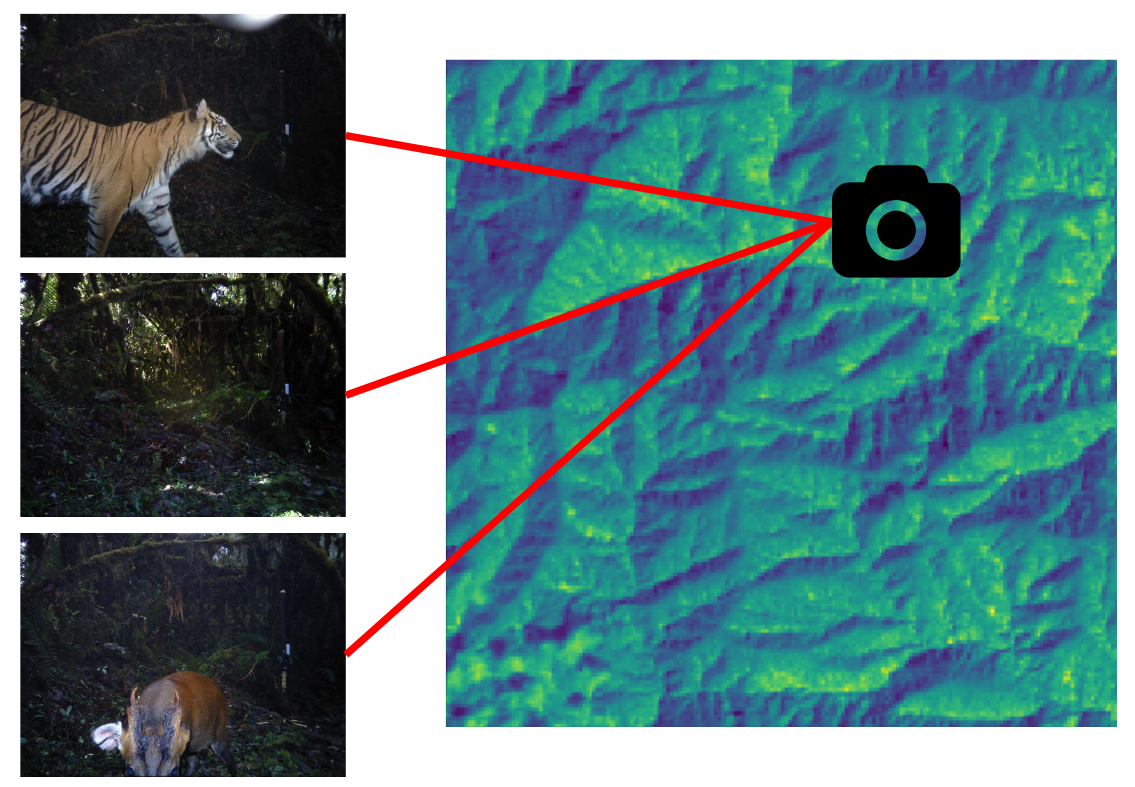}
    \caption{\textbf{The iWildCam 2020 dataset.} This year's dataset includes data from multiple modalities: camera traps, citizen scientists, and remote sensing. Here we can see an example of data from a camera trap paired with a visualization of the infrared channel of the paired remote sensing imagery.}
    \label{fig:splash}
\end{figure}

The 2020 iWildCam challenge includes a new component: the use of multiple data modalities (see Fig. \ref{fig:splash}).
An ecosystem can be monitored in a variety of ways (e.g. camera traps, citizen scientists, remote sensing) each of which has its own strengths and limitations.
To facilitate the exploration of techniques for combining these complementary data streams, we provide a time series of remote sensing imagery for each camera trap location as well as curated subsets of the iNaturalist competition datasets matching the species seen in the camera trap data.
It has been shown that species classification performance can be dramatically improved by using information beyond the image itself \cite{macaodha2019presence,chu2019geo,beery2019long} so we expect that participants will find creative and effective uses for this data.

\section{Data Preparation}
The dataset consists of three primary components: (i) camera trap images, (ii) citizen science images, and (iii) multispectral imagery for each camera location.
% EC: I made this a little more terse because the details will follow. Trying to remove some repetition. 
% On the competition GitHub page we provide the multispectral data, a taxonomy file mapping our classes into the iNat taxonomy, a subset of iNat data mapped into our class set, and a camera trap detection model (the MegaDetector) along with the corresponding detections.
 
\subsection{Camera Trap Data}
The camera trap data (along with expert annotations) is provided by the Wildlife Conservation Society (WCS) \cite{wcs_cam_traps}. 
We split the data by camera location, so no images from the test cameras are included in the training set to avoid overfitting to one set of backgrounds \cite{beery2018recognition}.

The training set contains $217,959$ images from $441$ locations, and the test set contains $62,894$ images from $111$ locations. 
These $552$ locations are spread across 12 countries in different parts of the world.
Each image is associated with a location ID so that images from the same location can be linked. 
As is typical for camera traps, approximately 50\% of the total number of images are empty (this varies per location).

\begin{figure}
\begin{minipage}[b]{\linewidth}
  \centering
  \centerline{\includegraphics[width=9cm]{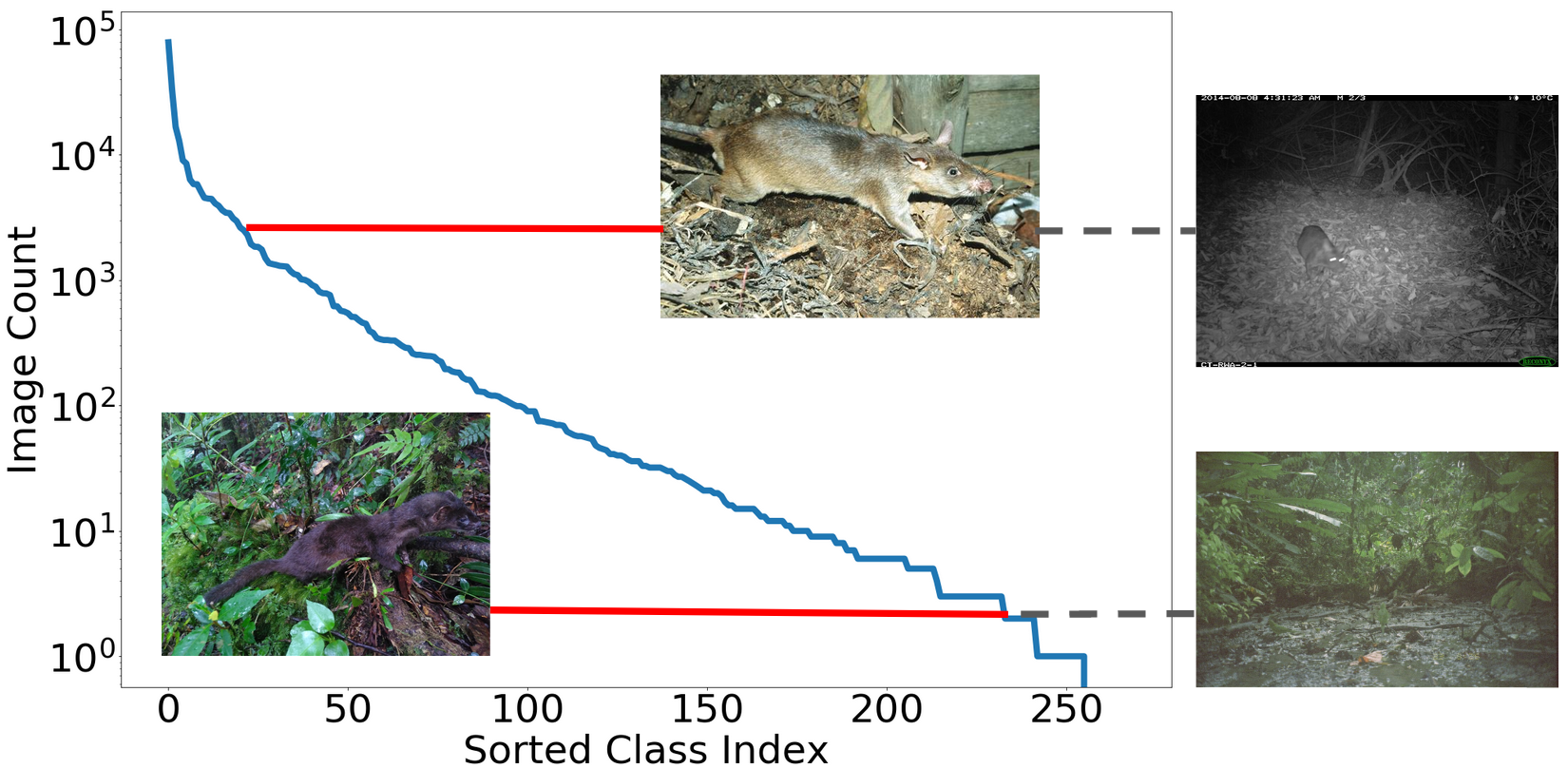}}
\end{minipage}

\caption{\textbf{Camera trap class distribution.} Per-class distribution of the camera trap data, which exhibits a long tail. We show examples of both a common class (the African giant pouched rat) and a rare class (the Indonesian mountain weasel). Within the plot we show images of each species, centered and focused, from iNaturalist. On the right we show images of each species within the frame of a camera trap, from WCS.}
\label{fig:camera_trap_distribution}
\end{figure}

% classes
There are 276 species represented in the camera trap images.
The class distribution is long-tailed, as shown in Fig. \ref{fig:camera_trap_distribution}. 
Since we have split the data by location, some classes appear only in the training set.
Any images with classes that appeared only in the test set were removed. 
%When using a trained model on a new camera location, some previously seen classes may not be present - 
%Similarly, in our dataset some classes appear only in the training set.
% This was allowed, as it emulates the absence of some expected species at unseen camera trap locations, a common challenge for real time wildlife annotation systems. 
% We chose to remove any classes that were seen only at the test locations, to focus this year's challenge on multimodality with remote sensing data. 
% iWildCam 2019 \cite{beery2019iwildcam} focused on open-set classification, with some test species not seen during training. 
% We plan to incorporate that realistic and difficult scenario again in future iWildCam competitions. 

\subsection{iNaturalist Data}
iNaturalist is an online community where citizen scientists post photos of plants and animals and collaboratively identify the species \cite{inat}. 
%in photos, examples of which can be seen in Fig. \ref{fig:inaturalist_comparison}. 
To facilitate the use of iNaturalist data, we provide a mapping from our classes into the iNaturalist taxonomy.\footnote{Note that for the purposes of the competition, competitors may only use iNaturalist data from the iNaturalist competition datasets.}
We also provide the subsets of the iNaturalist 2017-2019 competition datasets \cite{van2018inaturalist} that correspond to species seen in the camera trap data.
This data provides $13,051$ additional images for training, covering $75$ classes. 

\begin{figure}
\begin{minipage}[b]{\linewidth}
  \centering
  \centerline{\includegraphics[width=9cm]{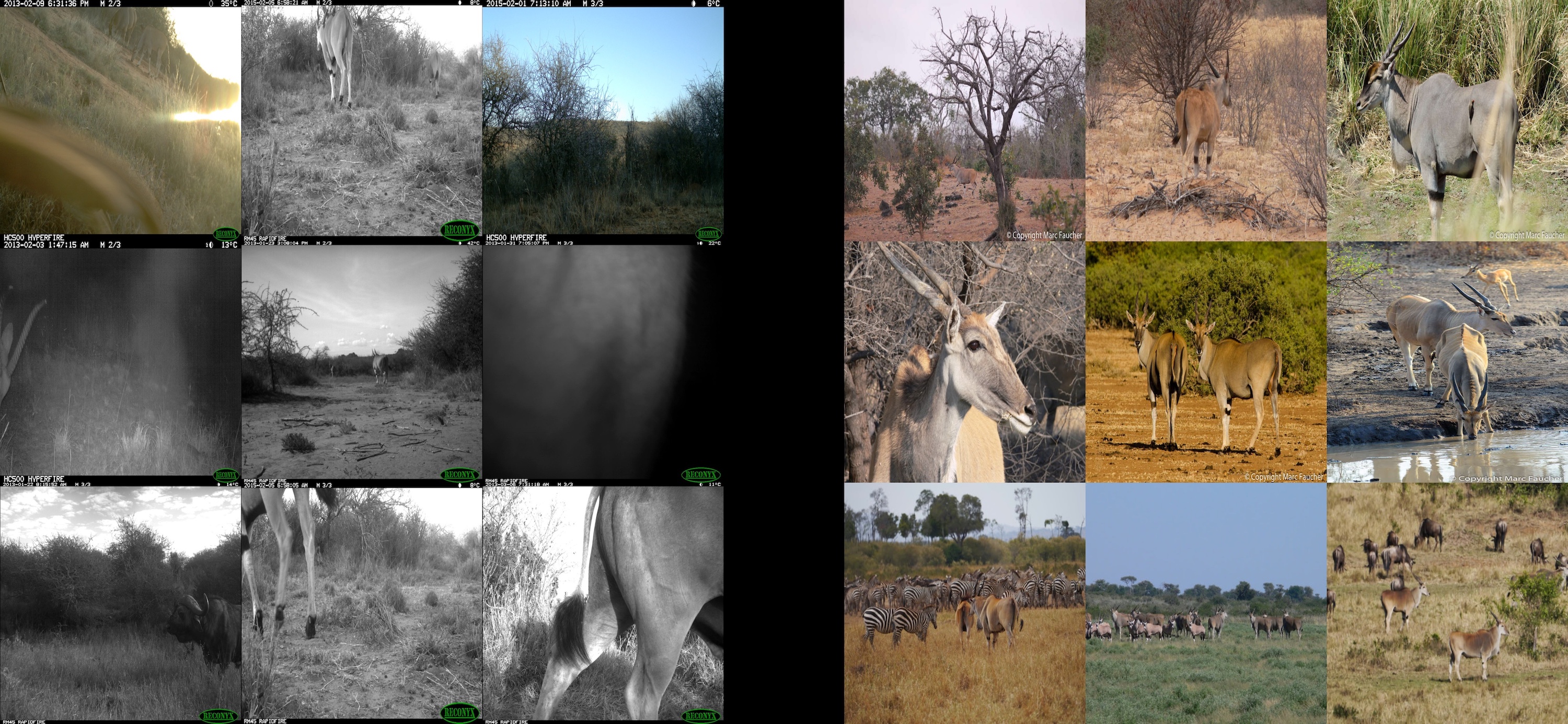}}
  \centerline{(1) Class ID 101}\medskip
\end{minipage}
\begin{minipage}[b]{\linewidth}
  \centering
  \centerline{\includegraphics[width=9cm]{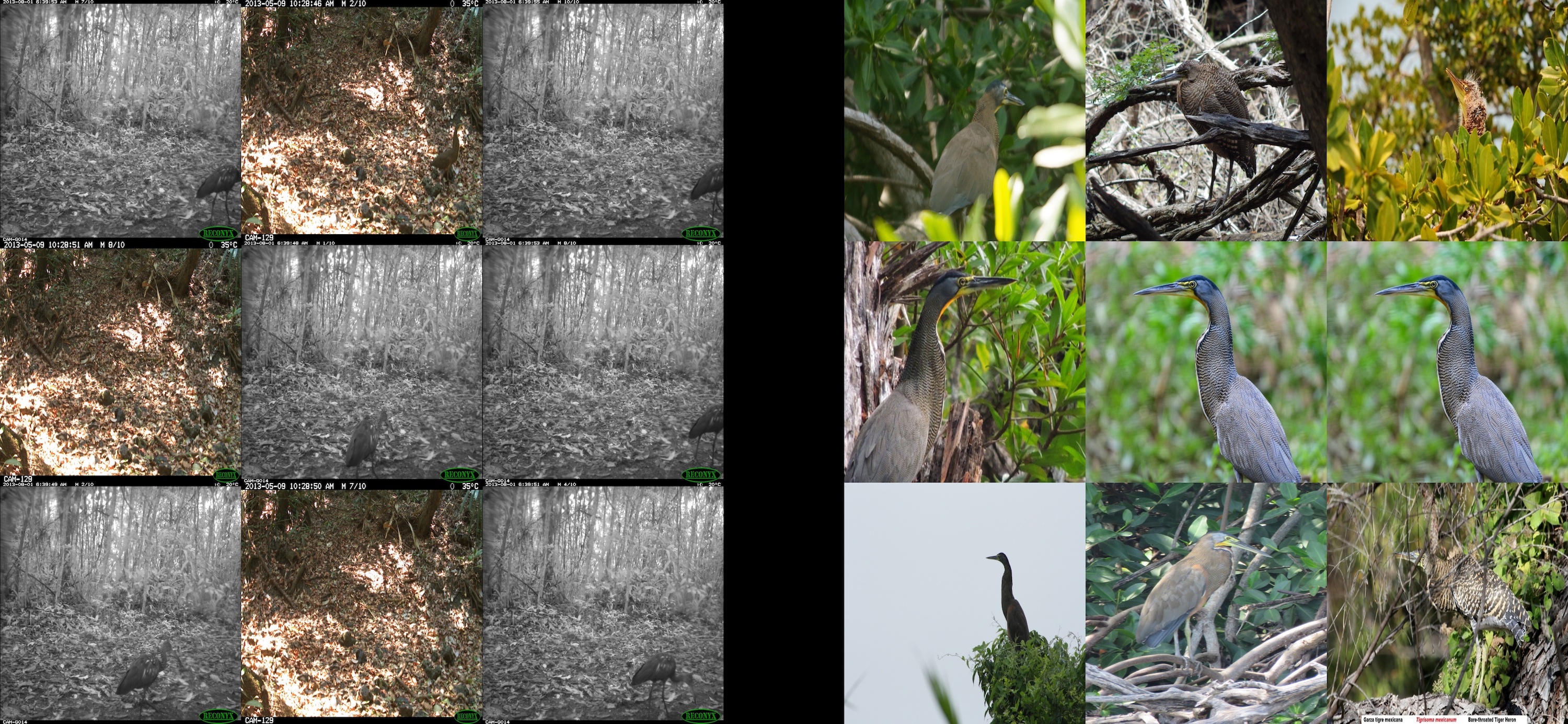}}
  \centerline{(2) Class ID 563}\medskip
\end{minipage}
\begin{minipage}[b]{\linewidth}
  \centering
  \centerline{\includegraphics[width=9cm]{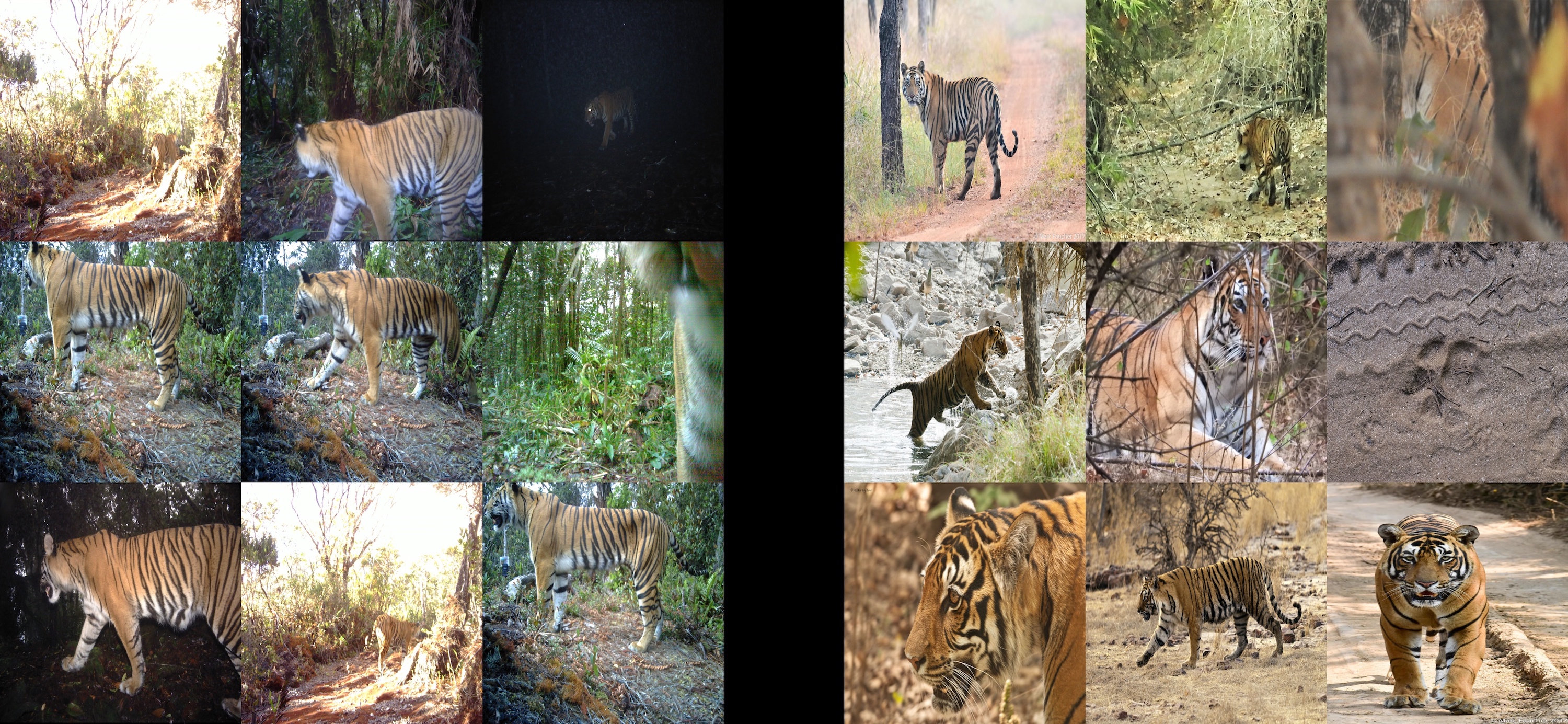}}
  \centerline{(3) Class ID 154}\medskip
\end{minipage}

\caption{\textbf{Camera trap data (left) vs iNaturalist data (right).} (1) Animal is large, so camera trap image does not fully capture it. (2) Animal is small, so it makes up a small part of the camera trap images. (3) Quality is equivalent, although iNaturalist images have more camera pose and animal pose variation.}
\label{fig:inaturalist_comparison}
\end{figure}

Though small relative to the camera trap data, the iNaturalist data has some unique characteristics.
First, the class distribution is completely different (though it is still long tailed).
Second, iNaturalist images are typically higher quality than the corresponding camera trap images, providing valuable examples for hard classes. 
See Fig. \ref{fig:inaturalist_comparison} for a comparison between iNaturalist images and camera trap images.
% which contains randomly sampled images for a few classes along with a few qualitative observations.

\subsection{Remote Sensing Data}
For each camera location we provide multispectral imagery collected by the Landsat 8 satellite \cite{landsat8}.
All data comes from the the Landsat 8 Tier 1 Surface Reflectance dataset \cite{t1sr} provided by Google Earth Engine \cite{gorelick2017google}.
This data has been been atmospherically corrected and meets certain radiometric and geometric quality standards.

\textbf{Data collection.}
The precise location of a camera trap is generally considered to be sensitive information, so we first obfuscate the coordinates of the camera. 
For each time point when imagery is available (the Landsat 8 satellite images the Earth once every 16 days), we extract a square \emph{patch} centered at the obfuscated coordinates consisting of 9 bands of multispectral imagery and 2 bands of per-pixel metadata. 
Each patch covers an area of 6km $\times$ 6km. 
Since one Landsat 8 pixel covers an area of 30m$^2$, each patch is $200 \times 200 \times 11$ pixels. 
Note that the bit depth of Landsat 8 data is 16. 

The multispectral imagery consists of 9 different bands, ordered by descending frequency / ascending wavelength.
Band 1 is ultra-blue.
Bands 2, 3, and 4 are traditional blue, green, and red. 
Band 5-9 are infrared.
Note that bands 8 and 9 are from a different sensor than bands 1-7 and have been upsampled from 100m$^2$/pixel to 30m$^2$/pixel.
Refer to \cite{t1sr} or \cite{landsat8} for more details.

Each patch of imagery has two corresponding \emph{quality assessment} (QA) bands which carry per-pixel metadata.
The first QA band (\texttt{pixelqa}) contains automatically generated labels for classes like \texttt{clear}, \texttt{water}, \texttt{cloud}, or \texttt{cloud shadow} which can help to interpret the pixel values.
The second QA band (\texttt{radsatqa}) labels the pixels in each band for which the sensor was saturated. 
Cloud cover and saturated pixels are common issues in remote sensing data, and the QA bands may provide some assistance.
However, they are automatically generated and cannot be trusted completely. 
See \cite{t1sr} for more details.

% \textbf{Data format.} 
% For camera location \texttt{XXX} there is a folder named \texttt{iwc\_XXX}. 
% The files in this folder are named \texttt{iwc\_XXX\_YYYY-MM-DD\_Z.npy}
% where \texttt{Z} is replaced by either \texttt{multispectral} ($200\times200\times9$), \texttt{pixelqa} ($200\times200\times1$), or \texttt{radsatqa} ($200\times200\times1$). 

\section{Baseline Results}
We trained a basic image classifier as a baseline for comparison.
% iNaturalist data and Remote Rensing data were not used in this model. 
The model is a randomly initialized Inception-v3 with input size $299 \times 299$, which was trained using only camera trap images.
During training, images were randomly cropped and perturbed in brightness, saturation, hue, and contrast.
We used the \texttt{rmsprop} optimizer with an initial learning rate of 0.0045 and a decay factor of 0.94. % Are there any other parameter values, even default ones, that we should include? 

Let $C$ be the number of classes. 
We trained using a class balanced loss from \cite{cui2019classbalanced}, given by 
\[ \mathcal{L}' (\mathbf{p}, y) = \frac{1 - \beta}{1 - \beta^{n_y}} \mathcal{L} (\mathbf{p}, y) \]
where $\mathbf{p} \in \mathbb{R}^C$ is the vector of predicted class probabilities (after softmax), $y \in \{1,\ldots,C\}$ is the ground truth class, $\mathcal{L}$ is the categorical cross-entropy loss, $n_y$ is the number of samples for class $y$, and $\beta$ is a hyperparameter which we set to 0.9. 

This baseline achieved a macro-averaged F1 score of $0.62$ and an accuracy of $62\%$ on the iWildCam 2020 test set.

% A flavor of class balancing was used according to \cite{cui2019classbalanced}, which weighted the loss function per class as 
% $$
% \mathcal{L}' (\mathbf{p}, y) = \frac{1 - \beta}{1 - \beta^{n_y}} \mathcal{L} (\mathbf{p}, y)
% $$
% where C is the total number of classes, $y \in \{1, ... C\}$ is the ground truth class, $\mathbf{p} \in \mathbf{R}^C$ is the vector of class probabilities, $\mathcal{L}$ is the loss function (softmax loss, in this case) and $\beta$ is a hyperparameter defined in \cite{cui2019classbalanced} (we used $\beta = 0.9$).

\section{Conclusion}
% Camera traps provide a unique experimental context which allows us to study generalization while controlling for many nuisance factors. 
The iWildCam 2020 dataset provides a test bed for studying generalization to new locations at a larger geographic scale than previous iWildCam competitions \cite{beery2018iwildcam,beery2019iwildcam}.
In addition, it facilitates exploration of multimodal approaches to camera trap image classification and pairs remote sensing imagery with camera trap imagery for the first time. 
% This dataset is the first to explicitly study the effect of added remote sensing data in the challenge of generalization to novel camera deployments. 

In subsequent years, we plan to extend the iWildCam challenge by adding additional data streams and tasks, such as detection and segmentation. We hope to use the knowledge we gain throughout these challenges to facilitate the development of systems that can accurately provide real-time species ID and counts in camera trap images at a global scale. Any forward progress made will have a direct impact on the scalability of biodiversity research geographically, temporally, and taxonomically.

\section{Acknowledgements}
We would like to thank Dan Morris and Siyu Yang (Microsoft AI for Earth) for their help curating the dataset, providing bounding boxes from the \href{https://github.com/microsoft/CameraTraps/blob/master/megadetector.md}{MegaDetector}, and hosting the data on Azure. We also thank the Wildlife Conservation Society for providing the camera trap data and annotations. We thank Kaggle for supporting the iWildCam competition for the past three years. Thanks also to the FGVC Workshop, Visipedia, and our advisor Pietro Perona for continued support. This work was supported in part by NSF GRFP Grant No. 1745301. The views are those of the authors and do not necessarily reflect the views of the NSF.

{\small
\bibliographystyle{ieee_fullname}
\bibliography{main}
}

% \appendix
% \section{Dataset Statistics}

% Eli TODOs:
% \begin{itemize}
%     \item Characterize number of time points per site.
%     \item Characterize time range coverage at different sites.
%     \item Characterize cloud cover and fill pixels.
% \end{itemize}

% \section{QA Band Format}

\end{document}